# Extractive Summarization of Call Transcripts

Pratik K. Biswas and Aleksandr Iakubovich

AI & Data Science, Global Network and Technology, Verizon Communications, New Jersey, USA

*Abstract*—Text summarization is the process of extracting the most important information from the text and presenting it concisely in fewer sentences. Call transcript is a text that involves textual description of a phone conversation between a customer (caller) and agent(s) (customer representatives). This paper presents an indigenously developed method that combines topic modeling and sentence selection with punctuation restoration in condensing ill-punctuated or un-punctuated call transcripts to produce summaries that are more readable. Extensive testing, evaluation and comparisons have demonstrated the efficacy of this summarizer for call transcript summarization.

## I. INTRODUCTION

In recent years, there is an abundance of multi-sourced information available for public consumption, fueled by the growth of the internet. In many cases, this volume of readily available text needs effective summarization to be useful for different purposes. It is very difficult for human beings to summarize large quantities of text manually. Hence automatic text summarization has become a very desirable tool in today's information age. It has the task of producing concise, fluent and readable summaries from larger bodies of text, while still preserving the original information content and meaning. Such summarization can be very useful, when applied to various domains, e.g., news articles, emails, call transcripts, medical history, mobile text messages, etc. Many such summarizers are available online on the internet like, Microsoft News2 & Google1 for news articles [40], MEAD & SWESUM for biomedical information [37], WikiSummarizer for Wikipedia articles, etc.

Numerous approaches have been developed for automatic text summarization and can be broadly classified into two groups: extractive summarization and abstractive summarization. Extractive summarization extracts important sentences from the original text and reproduce them verbatim in the summary, while abstractive summarization generates new sentences.

Call transcripts are written texts originally presented in a different medium and so call transcription is defined as the process of converting a voice or video call audio track into written words through speech to text conversion, to be stored as plain text in a conversational language. In this paper, however, we will be confining ourselves to textual descriptions of audio recordings of voice calls between customer (caller) and agent(s) (customer representatives) of a phone company. Automatic summarization of call transcripts, in our consideration, pose certain unique challenges, as follows: 1) they are not continuous texts but include conversation between customers and agents, 2) they are often very long and are embedded with "small talks" and can include a large number of sentences that are irrelevant and even meaningless, 3) they include several ill-formed, grammatically incorrect sentences, 4) they are either un-punctuated or are improperly punctuated based on *pauses* in the conversation as perceived by annotators and so are often unreadable, and 5) existing open-source summarization tools don't perform too well with call transcripts.

In this paper, we have presented a novel extractive summarization technique that combines *channel separation* (separation into customer and agent transcripts), *topic modeling, sentence selection* with *punctuation restoration* to produce properly punctuated, fixed-length and readable customer and agent summaries from the original call transcripts, that can adequately summarize customer concerns and agent resolutions.

## II. RELATED WORK

Related research can be broadly grouped into two categories: 1) extractive summarization and 2) abstractive summarization. Research under the first category is most relevant to our work.

Radef et al. [36] defined summary as "a text that is produced from one or more texts, that conveys important information in the original text(s), and that is no longer than half of the original text(s) and usually, significantly less than that." Automatic text summarization gained attraction as early as the 1950s. Different methods and extensive surveys of automatic text summarization have been provided in [1, 12, 13, 15, 16, 17, 27, 34, 39, 40, 49].

Luhn et al. [26] introduced a method to extract salient sentences from the text using features such as word and phrase frequency. They proposed to weight the sentences of a document as a function of high frequency words, ignoring very high frequency common words. Edmundson et al. [10] described a paradigm based on key phrases where they used four different methods to determine the sentence weight. Kupiec et al. [20] developed the Trainable Document Summarizer, which performed the sentence-extracting task, based on a number of weighting heuristics. Bookstein et al. [3] built clusters of index terms, phrases and other sub-parts of documents for extractive text summarization. Brandow et al. [4] launched the ANES text extraction system that automatically condensed domain-independent electronic news data. Conroy et al. [7] & Mittendorf et al. [29] used hidden markov models for text summarization. Chen et al. [6]

discussed a sentence selection based approach to text summarization, while Gong et al. [14] and Wang et al. [46] described how multiple documents could be summarized using topic models. Neto et al. [33] introduced machine learning approaches to automatic text summarization and Kaikhah [19] discussed how neural networks could be useful for summarizing news articles. Suanmali et al. [43] proposed fuzzy logic based extractive text summarization to improve the quality of the summaries created by the general statistical method. Nallapati et al. [30] presented a Recurrent Neural Network (RNN) based sequence model for extractive summarization of documents. Narayan et al. [32] conceptualized extractive summarization as a sentence-ranking task and proposed a novel training algorithm for optimizing the ROUGE evaluation metric [22] through a reinforcement learning objective. Xu et al. [48] constructed a neural model for single-document summarization based jointly on extraction and syntactic compression. Verma et al. [45] applied a Restricted Boltzmann Machine to enhance the summaries of factual reports created through extractive summarization. Miller [28] used Bidirectional Encoder Representations from Transformers (BERT) for summarization of lecture notes. Liu [24] described BERTSUM, a simple variant of BERT, for extractive summarization. Liu et al. [25] displayed how BERT could be generally applied in text summarization and proposed a general framework for both extractive and abstractive models. Zhong et al. [50] formulated the extractive summarization task as a semantic text-matching problem, while Wu et al. [47] instituted a new text-to-graph task of predicting summarized knowledge graphs from long documents. Lemberger et al. [21] recounted the applicability of deep learning models for automatic text summarization.

Lin et al. [23] surveyed the state of the art in abstractive summarization, while Khan et al. [18] reviewed various abstractive summarization methods. Nallapati et al. [31], Paulus et al. [35] and See et al. [41] employed recurrent neural network, deep reinforcement learning and pointer-generator network for abstractive summarization.

### III. MAJOR CONTRIBUTIONS

In this paper, we have proposed an innovative extractive summarization technique for call transcript summarization. Our main contributions and advantages can be summarized as follows:

- We have integrated topic modeling and embedding based sentence selection with transformer based punctuation restoration for extractive summarization through a novel 10-step sequential procedure (method).
- Our method splits the original call transcript into customer and agent transcripts by the associated channel identifiers and then summarizes each transcript separately for more coherent results.
- Our method restores full punctuation in the summaries of un-punctuated or ill-punctuated call transcripts.
- We have uniquely modified and retrained the BERT transformer model architecture for punctuation restoration by adding a classification layer above BERT's 12 layers.
- Our method creates, compares and evaluates the performances of multiple different types of topic models for the transcripts, before selecting the most optimal one for extractive summarization; it also provides the option to specify the topic model type to be used for extractive summarization and allows the summarizer to use different topic model types to generate customer and agent summaries.
- We have introduced a new metric for measuring the effectiveness of punctuation restoration in the punctuated summaries.

### IV. PRELIMINARIES – CONCEPTS AND TERMINOLOGIES

In this section, we clarify key concepts and terminologies and explain certain technologies, which provide the foundation for our work.

The automatic summarization of text is a well-defined task in the field of *Natural Language processing (NLP)*. *Automatic text summarization* attempts to convert a larger document into a shorter version preserving its information content and overall meaning. A good summary should reflect the diverse topics of the document while keeping redundancy to a minimum [15]. In general, there are two different approaches to automatic text summarization, namely, extractive and abstractive.

*Extractive summarization* methods identify the relevant sections in the original text, extract the most important paragraphs, sentences, phrases, etc. from there and concatenate them into shorter form. In contrast, *abstractive summarization* methods attempt to convey the most important information from the original text by generating new sentences. In other words, they interpret, examine and analyze the original text using advanced natural language techniques to get a better understanding of the content and then describe it through shorter and more focused text, comprising of new sentences. Purely extractive summaries often give better results than automatic abstractive summaries [11]. This is because of the fact that abstractive summarization methods cope with problems like semantic representation, inference and natural language generation, which are relatively harder than data-driven approaches such as sentence extraction [1]. Most abstractive summarization techniques, specifically the ones using deep learning, also depend upon extractive summarization to extract the summaries for the training samples from which they train to generate new text. In this paper, we focus only on extractive summarization, as it is relevant to our work.

#### A. Extractive Summarization

*Extractive summarization* techniques generate summaries by selecting the most important sentences, paragraphs, etc., from the original text. The importance of sentences is decided based on statistical and linguistic features of sentences [18]. Input can be either single or multiple documents or sources of text. Extractive summarization consists of three main steps, namely, intermediate representation of input text, scoring of sentences based on the intermediate representation, selection of sentences for summary generation. There are two approaches, namely, topic based and indicator based, which are used for intermediate representation of original text. Topic representation based approaches transform input text into constituent topics. They are further subdivided into frequency-driven, topic word based, cluster based, latent semantic analysis dependent and Bayesian topic model based methods. Indicator representation based approaches characterize sentences in the input text through features such as sentence length, position in the document, having certain phrases, etc. They can be further subdivided into graph-theoretic, fuzzy-logic driven, machine learning based and neural network based methods.

*B. Latent Semantic Analysis*

*Latent semantic analysis (LSA)*, also known as *Latent Semantic Indexing (LSI)*, is an unsupervised method for extracting a representation of text semantics based on observed words. It tries to bring out latent relationships within a collection of documents on to a lower dimensional space. LSA is based on the principle that words that are close in meaning will occur in similar pieces of text (the distributional hypothesis). It uses a mathematical technique called *singular value decomposition (SVD)* to identify patterns in the relationships between the terms and concepts contained in unstructured texts. It was introduced by Deerwester et al. in [8]. Gong et al. [14] proposed a LSA-based method to select highly ranked sentences for single and multi-document news summarization.

*C. Bayesian Topic Models*

*Topic modeling* can be described as a type of a statistical method for finding a group of words (i.e., topic), from a collection of documents, that best represents the information in the collection. Bayesian topic models are *unsupervised probabilistic models* that uncover and represent the topics of documents or source texts [42]. They have gained huge popularity in recent years. Their advantage in describing and representing topics in detail enables the development of summarizer systems, which can use them to determine the similarities and differences between documents to be used in summarization [27].

There are many techniques, which are used to obtain probabilistic topic models. *Latent Dirichlet Allocation (LDA)* is one such widely used topic modelling technique that represents the documents as a random mixture of latent topics, where each topic is a probability distribution of words [2]. It has been used in recent times for multi-document summarization. Wang et al. [46] introduced a Bayesian sentence-based topic model for summarization, which using both term-document and term-sentence associations, achieved significant performance improvement and outperformed many other summarization methods. *Hierarchical Dirichlet Process* (HDP) is another topic modeling technique, which is an extension of LDA. It is a nonparametric Bayesian approach, which uses a mixed-membership model for unsupervised analysis of grouped data. Unlike LDA (its's finite counterpart), HDP infers the number of topics from the data.

*D. Transformers*

*Transformers* in NLP provide general-purpose architectures for *Natural Language Understanding (NLU)* and *Natural Language Generation (NLG)* with over 32+ pre-trained models. They were first introduced in [44]. Transformers are *Seq2Seq deep learning* models that transform sequential inputs to sequential outputs. However, they are based solely on *attention* mechanisms, dispensing entirely with *recurrence* and *convolutions* of the earlier deep learning architectures. Transformers do not require that the sequential data be processed in order, which allows for much more parallelization than *Recurrent Neural Networks (RNNs)* and therefore reduced training times [44]. Since their introduction, transformers have become the model of choice for tackling many problems in NLP, replacing older recurrent neural network models such as the *Long Short-term Memory (LSTM)*. Transformer models can train on much larger datasets than before, as they can support more parallelization during training. This has resulted in the development of pre-trained systems such as *Bidirectional Encoder Representations from Transformers (BERT)*, which have been trained with huge general language datasets, and can be fine-tuned to specific linguistic tasks [9]. BERT is a bidirectional transformer pre-trained using a combination of *masked language* modeling objective and next sentence prediction on a large corpus comprising the Toronto Book Corpus and Wikipedia, by *jointly conditioning on both left and right contexts in all layers. Consequently, a pre-trained BERT model can be fine-tuned with just one additional output layer to create state-of-the-art models for a wide range of NLP tasks*.

Transformers employ a 12-layered *encoder-decoder* architecture that comprises a stack of 6 encoding layers that processes the input iteratively one layer after another and another stack of 6 decoding layers that does the same thing to the output of the encoder. The encoders are all identical in structure. Each one is broken down into two sub-layers, namely, *self-attention* and *feed-forward neural network*. The decoder has one more layer between them, which is an *attention* layer that helps it to focus on relevant parts of the input sentence (similar to what attention does in *Seq2Seq models*). Therefore, when we pass a sentence into a transformer, it is embedded and passed into a stack of encoders. The output from the final encoder is passed into each decoder block in the decoder stack, which then generates the output.

*E. Embeddings*

*Embeddings* are mathematical functions that map "entities" to a latent space with complex and meaningful dimensions. Words or sentences or paragraphs can be mapped into a shared latent space such that the meaning of the word/sentence/paragraph can be represented geometrically. Machine learning approaches towards NLP require words to be expressed in vector form. Word embeddings, proposed in [38], is a feature engineering technique in which words are mapped into a vector of real numbers in a pre-defined vector space. It is a learned representation for text where words that have the same meaning have a similar representation. The idea of using a dense distributed representation for each word is a key to the approach. *Word2Vec, GloVe*, etc. provide pre-trained word embedding models in a type of *transfer learning*. Embedding techniques initially focused on *words* but the attention soon started to shift to other types of textual content, such as *n-grams*, *sentences* and *documents*. The *Universal Sentence Encoder (USE)* encodes text into high dimensional vectors that can be used for text classification, semantic similarity, clustering and other natural language tasks [5]. The model is trained and optimized for sentences, phrases or short paragraphs, from a variety of data sources with the aim of dynamically accommodating a wide variety of natural language understanding tasks. The model maps variable length input English text into an output of a 512 dimensional vector.

V. EXTRACTIVE SUMMARIZATION OF CALL TRANSCRIPTS

This section provides a description of an extractive summarization technique that we are proposing for the summarization of call transcripts. This extractive summarization technique uniquely integrates *channel (speaker) separation*, *topic modeling*, and *similarity based sentence selection* with *punctuation restoration* through a 10-step sequential procedure/method. The procedure is highly parameterized. The following are its ten self-contained steps, each with its brief description.

1. Call Transcript Channel (Speaker) Separation – Separate each call transcript into customer and agent transcripts based on its channel or speaker identifier, by iterating through all transcripts.
2. Partial Punctuation Restoration – Preprocess transcripts (customer & agent) to remove existing punctuations and use a *transformer* based model to restore punctuations *partially*, i.e., restore only *periods* as delimiters, so that sentences can be separated in each transcript; by iterating through all transcripts.
3. Document Preparation – Preprocess transcripts and generate *documents* from customer and agent transcripts, by iterating through all transcripts, i.e., one document from each transcript, where each document is a list of words from that transcript, obtained through NLP pipeline based preprocessing.
4. Topic Modeling – Build and optimize *different types* of *topic models* using the *vocabularies*, *corpus* and *documents* from all of customer and agent transcripts, and then pick the best customer and agent topic models based on their coherence scores.
    i. Build different types of customer & agent topic models, e.g., *LDA*, *LSI*, *HDP* by varying their hyper-parameter (e.g., topic number) values within pre-specified ranges and evaluate the models using their coherence scores ($c\_v$, $u\_mass$, etc.).
    ii. Select the most optimal (or near about) topic models for customer and agent transcripts.
    The topic model type is a parameter of this procedure and so if it is provided during invocation then model optimization is confined to only that topic model type in step 4-i for best model selection.
5. Dominant Topic Identification – Get the most *dominant topic(s)* from the aforesaid topic models with the associated keywords for each of customer and agent documents in every pair, by iterating through all transcripts.
6. Significant Term Selection – Get the most *relevant keywords/terms* from each pair of customer and agent transcripts by doing *term-based similarity analysis* between the keywords of the corresponding dominant topics, using one of the following two approaches, by iterating through all transcripts.
    i. Global Extraction – Extract *terms* from the keywords associated with each pair of dominant topics, which need not be necessarily, present in the transcripts (customer & agent) themselves.
    ii. Local Extraction – Extract *terms* from (local to) the customer and agent transcripts that are similar to the corresponding dominant topic keywords and are also similar to themselves.
    The choice for the term extraction method has also been parameterized for the procedure.
7. Summary Generation – Generate fixed-length (user-specified number) customer and agent transcript summaries, by iterating through all pairs of customer and agent transcripts.
    i. Identify the *most unique sentences* in each of customer and agent transcripts in every pair, if necessary, based on *similarity analysis* among all sentences of the corresponding transcript using *embeddings*.
    ii. Extract a fixed number (user-specified) of most relevant sentences from each of customer and agent transcripts through *sentence-based similarity analysis* between every sentence of the corresponding transcript and the string/document created out of the most significant terms for that pair of transcripts (step 6), using *embeddings*.
    The desired length of the summary (number of sentences) is a parameter of the procedure.
8. Punctuation Restoration – Remove *existing periods* from each pair of customer and agent summaries, restore *partial* and *full punctuation* using a *transformer* based model and post-process to make them more readable, by iterating through all of them.
9. Summary Tabulation – Save summaries of all transcripts in a table for future use.
10. Summarization Efficacy Determination – Evaluate summaries on *content* (information) and *readability* (punctuation restoration), by iterating through every pair of transcripts & their corresponding summaries.
    i. Summary Evaluation – Evaluate the *goodness* of summarization by *comparing* customer and agent summaries against original transcripts (or manually generated summaries) to generate average *rouge* scores.
    ii. Punctuation Restoration Evaluation – Evaluate the *correctness* of punctuation restoration by *matching* the number of punctuation symbols (*periods*) between the extracted and their partially

punctuated summaries for both customer and agent, to generate the average *accuracy* scores.

The *full punctuation restored summaries*, from step 8, are the outputs from this procedure. Next, we take a deeper look at some key steps of this procedure, discuss their implementations in detail, and provide algorithms where necessary.

*A. Channel Separation*

Call transcripts include conversations/dialogs between customer and one or more agents and so the resultant summaries can often get mixed up. Separation of a transcript into customer and agent transcripts can make each summary more coherent. Customer summaries can give better ideas of the problems, while the agent summaries can give a better understanding of the causes or the solutions.

Call transcripts are generally available to us as *json-formatted* objects. Hence, channel separation involves extraction of the transcript string from the *json-formatted* object, channel identification and then decomposition of the transcript into customer and agent transcripts using the associated channel identifiers. If the channel identifiers do not clearly identify the speakers then we can use a *pre-trained BERT transformer* model with a *linear classifier* from *PyTorch nn* module as an additional layer, on top of BERT's 12 layers, to classify each dialog of the transcript into one of the 2 classes, i.e., *customer* and *agent* and then combine each type of dialogs to create customer and agent transcripts.

*B. Document Preparation*

A *document* is a list of *keywords* extracted from each transcript and is used as input to the topic model. For document preparation, we have built a custom NLP preprocessing pipeline comprising of tokenization; punctuation, extended stop-words & small words (length ≤ 4) removal; regular expression matching; lowercasing; contraction mapping; bigrams and trigrams creation; lemmatization; parts of speech tagging & allowable tag selection. This has been implemented by combining modules (functionalities) available from 4 *Python* packages, namely, *re*, *spaCy*, *NLTK* and *genism*.

*C. Topic Model Optimization & Optimal Model Selection*

If the *topic model type* is specified at the invocation of the procedure, then we create multiple *topic models* of the desired type, for both customer and agent, using the *documents*, *corpus* and *vocabulary* from the corresponding call transcripts, by varying the hyper-parameter (e.g., topic number) values within the pre-defined ranges by the pre-defined steps; compute their coherence scores and identify the topic models and associated hyper-parameter values that produce the best scores.

Otherwise, by default, we perform the above-mentioned activity for all 3 different topic model types, namely, *LDA*, *LSI* and *HDP*, in parallel, and identify the topic models and associated hyper-parameter values that produce the best scores amongst topic models of all 3 types.

Fig. 1 displays the steps of this algorithm. For topic modeling, we have exclusively used the *Python* based *genism* package.

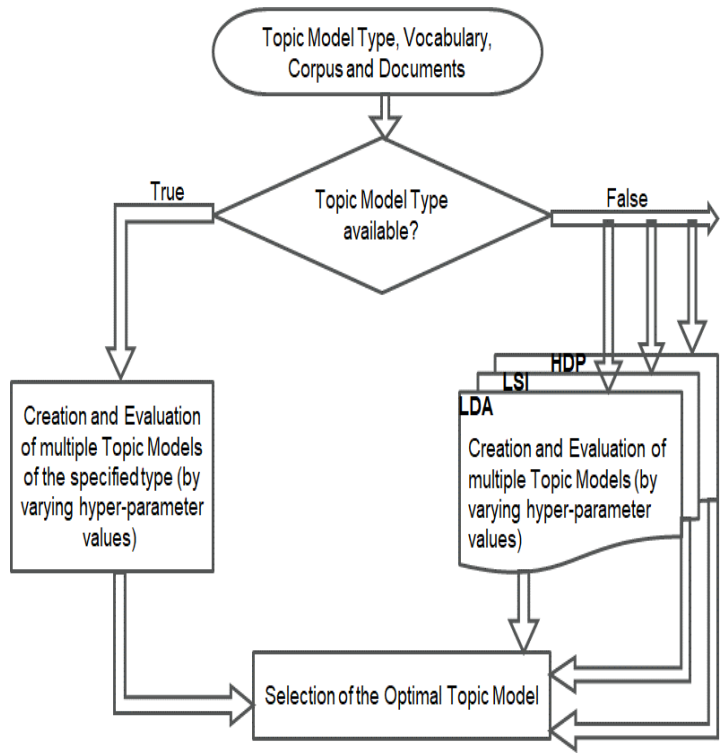

Fig. 1: Topic Model Optimization & Selection

*D. Punctuation Restoration*

Here, we describe the punctuation restoration algorithm, used in steps 2 and 8 of the previously mentioned proposed procedure, in detail.

We have used the *BertForMaskedLM* class of the *PyTorch BERT* model for punctuation restoration and added an additional *linear* layer (*PyTorch nn module*) above the 12 BERT layers. The output of original BERT layers is a vector with the size of all vocabulary. The additional linear layer takes this as input and gives as output one of four classes, i.e., "O" (Other), "Comma", "Period" and "Question" for each encoded word. We retrained this modified BERT model using *TED transcripts*, consisting of two million words. Different variations of punctuation restoration with BERT model have been presented earlier but the retraining with the proposed architecture is a unique approach for punctuation restoration.

The steps of this algorithm are as follows:
1) Preprocess either *transcript* to remove *duplicate words*, *phrases* and *expressions* or *punctuations* inserted as *delimiters* based on annotator's perceptions of the pauses in the conversation or a *summary* to remove *periods*. The output from this step is a continuous string representing the cleaned & unpunctuated text of the transcript/summary.
2) Instantiate the *pre-trained BERT punctuation model* and initialize it on GPUs to classify each encoded word in text to one of 4 classes, namely, 'Other': 0, 'Comma': 1, 'Period': 2 and 'Question': 3.
3) Tokenize transcript/summary and encode *tokens* to numeric format (*token identifier/ID*) using the *BertTokenizer*.

4) Create *segments* of surrounding token IDs for each encoded word (token ID) from the text and insert '0' as a *placeholder* halfway through each segment. The segment size is a parameter of the algorithm.
5) Use the placeholders from the above step to predict *punctuation class identifiers* (IDs) for all token IDs in the segments through the *modified BERT* model *classifications*.
6) Map class IDs to words/symbols and merge words (if needed) to restore 2 sets of *punctuated* transcripts/summaries, one with just *periods* (partial punctuation restoration) and the other with all punctuations (full punctuation restoration). *Partial punctuation restoration* is used in step 2, while both *partial* and *full punctuation restorations* are used in step 8 of the main procedure.

Fig. 2 displays the algorithm through block diagrams. We have found that the BERT model for punctuation restoration gives 30% more accurate results than the LSTM based model. We have implemented the punctuation restoration algorithm using *BERT Transformer*, *BertPunc* and *nn* modules, available from *PyTorch*.

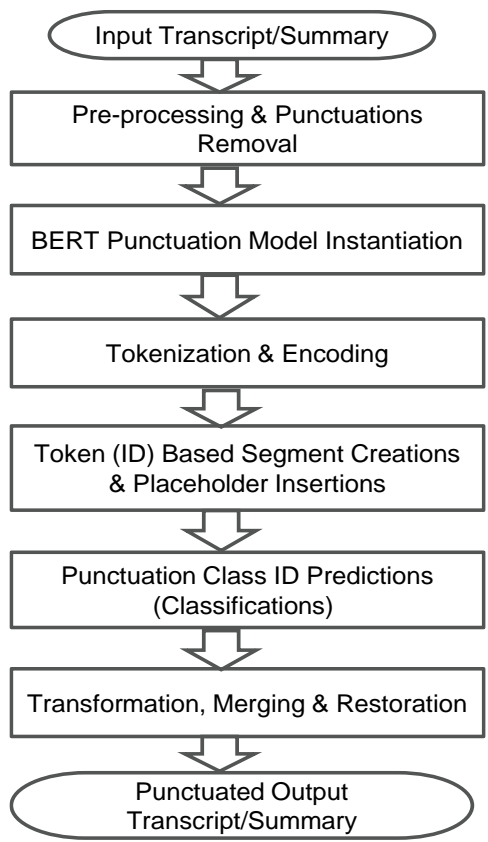

Fig.2: Punctuation Restoration

### E. Summary Generation through Sentence Selection

Next, we present the algorithm for generating separate customer and agent summaries from each pair of transcripts through *sentence selection*, starting with their corresponding *topic models*. In other words, the following algorithm implements steps 5 to 7 of the proposed procedure. The inputs to the algorithm are a pair of customer and agent transcripts and the corresponding optimized *topic models*, *corpus* & *documents* from all transcripts.

1) Use selected topic models (customer and agent) to identify *dominant topic(s)* from each of customer & agent documents for every pair and produce two lists of associated *keywords*, one for each of customer & agent transcripts. The number of dominant topics to be identified per transcript is a parameter of the procedure.
2) Use the keywords/terms associated with customer and agent dominant topics to extract the most significant inter-related *terms* for each pair of transcripts. This has been achieved using *word-based similarity analysis*. As mentioned before in the main procedure, we provide 2 alternatives to keyword/term extraction, where the choice is parameterized.

   If a *global extraction* is desired then use the two lists of keywords, associated with the corresponding customer and agent dominant topics in every pair, and identify *terms* that are most *similar* to each other, i.e., where the degree of similarity is above a certain parameterized threshold.

   Otherwise, if a *local extraction* is desired, then *first* find a set of *terms*, from each of customer and agent documents in every pair, that is most *similar* (above parameterized threshold) to the *keywords* associated with the corresponding *dominant topic*(s) for that transcript; *secondly* identify *terms* from these two sets of terms, extracted locally from customer and agent documents, that are most *inter-related*, i.e., where degree of similarity is above the parameterized threshold.
3) Construct a *string* (or *document*) with the *extracted significant terms* for each pair of customer and agent transcripts.
4) Identify the most *unique* sentences in each of the customer and agent transcripts in every pair using *embeddings* and eliminate the redundant sentences to condense the original transcripts. This is achieved by generating a *correlation matrix* with the *embeddings* for all the *sentences* in the original transcript and removing those sentences whose correlations are above a certain pre-specified threshold. The uniqueness threshold for sentence elimination is a parameter to the procedure.
5) Select a certain specified number (parameter to the procedure) of most *important* sentences from each of the condensed customer and agent transcripts in every pair that are *most similar* to the string constructed @ step 3 of the current procedure using *sentence-based embeddings*; list and concatenate them in order; subsequently present the results as the summaries for the corresponding transcripts.

Fig. 3 illustrates the crux of this algorithm for any given pair of transcripts through block diagrams. For *term-based similarity analysis*, we have used *Word2Vec* algorithm based *spaCy*'s *en_vectors_web_lg*; while for *sentence-based similarity analysis and summary generation*, we have used the *Universal Sentence Encoder (USE)* from *tensorflow-hub*, along with the *Python* based *pandas* and *numpy* packages.

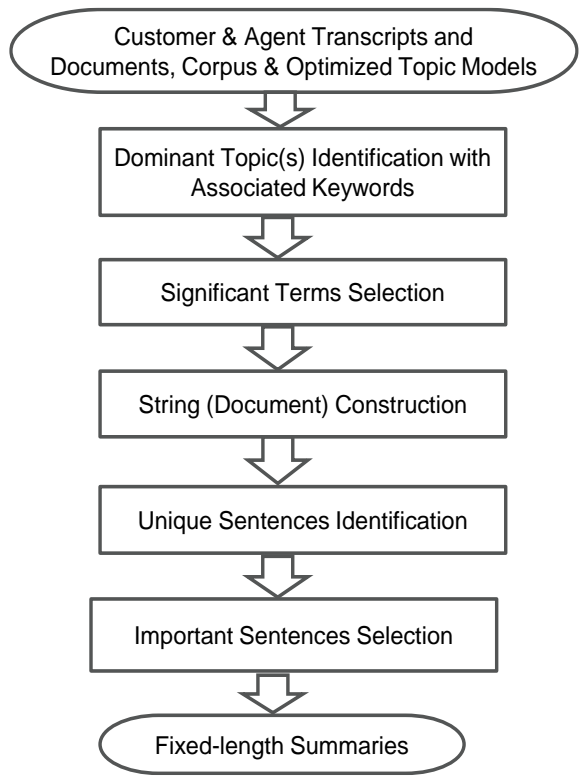

Fig. 3: Summary Generation through Sentence Selection

*F. Summarization Evaluation*

We have determined the effectiveness of the summarizer by measuring both the *goodness* (*quality*) of summarization and the *correctness* (*accuracy*) of the punctuation restoration reflecting the readability of the summaries.

For the *quality* of the information content of the generated summaries, we have used the metric *rouge* or *bleu* scores as a measurement of their *goodness*. We have compared the customer and agent summaries against the corresponding transcripts (or manually generated summaries if available) and computed their individual *rouge-l* or *bleu* scores using the *Python* packages *rouge* (*Rouge*) and *NLTK* (*nltk.translate.bleu_score*).

For the *correctness* of the punctuation restoration, we have defined the following metric named as *punctuation-restoration-accuracy score* to measure the *accuracy* of the algorithm.

***Definition 5.6.1***: *Punctuation-restoration-accuracy score represents the number of matches of punctuation symbols (periods) between the original/extracted text (transcript/summary) and the punctuated text (transcript/summary), expressed as a percentage (%).*

We have used the *accuracy_score* function from python's *sklearn.metrics* package to implement this metric. We have evaluated the effectiveness of the punctuation restoration algorithm through the following two steps.

1) Extract *periods* from both the extracted customer and agent summaries (step 7 of the main procedure) together with their *period-only* summaries (step 8 of the same procedure) and produce 4 lists, one each for *extracted* and *partially punctuated* summaries for each of customer and agent.
2) Count the number of *matches* between the two extracted lists of *periods* from the summaries of both customer & agent, and compute the above-defined *accuracy scores* for both. In each case the number reflects the % of *periods* from the partially punctuated (*periods-only*) summary found in the extracted summary.

We have computed the customer and agent *rouge* & *punctuation-restoration-accuracy* scores for every pair of customer & agent summaries, by iterating through all transcripts, & calculated their *averages* from their respective individual scores.

VI. PERFORMANCE EVALUATION

User satisfaction, effectiveness (quality of summaries & correctness of punctuations), efficiency (summarization time), flexibility, and performance comparison with another open-source, off-the-shelf extractive summarizer are some of the considerations that helped us evaluate the performance of our summarizer for call transcript summarization.

*A. Experimental Setup*

We set up a Spark cluster, consisting of a driver node and dynamically allocated multiple executor nodes for data collection, preprocessing and summarization. The NVIDIA CUDA Deep Neural Network (cuDNN v7.6) accelerated our training process for punctuation restoration. We re-trained and tested the modified BERT transformer model on NVIDIA Tesla V100-SXM2-32GB GPU based nodes. The driver node used anywhere between 1 to 4 GPUs. We have tested our extractive summarizer on 4 separate samples from 4 different use cases with 3 of these samples consisting of around 50 call transcripts and another larger one consisting of 1000 call transcripts. The evaluation of results have been both manual and automated.

*B. Manual Evaluation*

The summaries generated by the proposed method have been manually verified and validated for content and readability by 4 different user groups for the four different use cases spread across multiple business units. The goal was to see if the summaries were deemed generally useful for the purposes they were used in the specific use cases. The process was informal in nature and the evaluation was subjective. We relied on user (our customer) feedback and let them manage and control their own evaluation process and satisfaction levels. Four different user groups (customers), three internal and one external, have now manually evaluated close to 100 pairs of customer and agent summaries for the four different use cases.

For user evaluations, we were mainly looking for answers to questions, which were both generic and specific to the use cases. The following are some examples of the two kinds.

Generic
- Q1: Did the customer and agent summaries in general give a fair description of the main problems (complaints) and the resolutions, based on the original, unseparated call transcripts? If so then what percentage of customer and agent summaries accurately summarized the content?
- Q2: Did the summaries help users better comprehend the information content of the transcripts than the original transcripts themselves?
- Q3: Did the summaries capture other secondary issues (topics) besides the main issue (topic)? If so then how many?
- Q4: Did the punctuations help in making the summaries more readable for understanding the content of the transcripts?
- Q5: Have the punctuations been generally restored correctly?
- Q6: How did our summaries compare with manually generated summaries, if available? Note a manually generated summary would consist of a fixed number (same as that for the automated) of ordered sentences extracted manually from the "period-only" customer and agent transcripts generated at step 2 of the proposed procedure that the user would deem as most important from those transcripts.

Use Case Specific
- Q1: Would the user be able to send the agent summaries as short text messages to the customers to prevent them from making repeat calls?
- Q2: Would the summaries indicate the possibility of churning for callers, which are classified as churners? What percentage of customer summaries included *negative snippets* for churn classification?
- Q3: How did the summaries generated by the proposed 10-step approach compare with summaries generated by several other open-source, off-the-shelf summarizers from transcripts which were originally ill-punctuated (with periods) and where these transcripts (for external summarizers) didn't go through an accurate period restoration step, i.e., step 2, of the proposed procedure.

We have used the feedbacks from each use case to improve our method and the results. We are happy to report that different business units are now using our summaries for very different business purposes on an ongoing basis.

### C. Automated Evaluation

For automated evaluation, we have looked at *effectiveness* and *efficiency*. For measuring the *effectiveness* of our summarization and for comparing performances with another popular, open-source extractive summarizer, we have used the metric *rouge-l score*. We have determined the *efficacy* of our *punctuation restoration* algorithm using our own *punctuation-restoration-accuracy* score metric.

The *efficiency* of a summarizer is important to real world applications. We have measured the *efficiency* of our summarizer by recording the time taken by each of the 10 steps of our proposed procedure/method. We have also compared the *efficiency* of our *summary generation through sentence extraction algorithm* (step 7) with that of the same open-source extractive summarizer by recording the time taken by each to summarize each of the four different samples.

*1) Results and Summarizer Comparisons*

Fig. 4 and Fig. 5 ((a) & (b)) show short (~5 sentences) *summaries* from two call transcripts, after channel separations, describing customer complaints about internet and phone service disruptions and agents' confirmations about impending technicians' visits. Table I compares the *effectiveness & efficiency* of the proposed summarizer for shorter summaries with those from the open-source *BERT Extractive Summarizer* [28, 51], using the four different samples, on two different *evaluation metrics*. BERT Extractive Summarizer generated the summaries using the "period-restored" customer and agent transcripts from step 2 of the proposed procedure. It's "*ratio*" parameter was automatically adjusted, using the number of words in the transcript, to ensure that its summaries were of comparable (shorter) lengths. This is important as we found that *longer* the summary, the *more similar* it is to the original transcript and so *higher* the rouge score. For the larger sample sizes, where manual summaries were not available, we compared the generated summaries with the "period-restored" original transcripts (from step 2 of our procedure) for computing their corresponding *rouge scores*. This was done for the summaries from both the proposed method as well as the BERT Extractive Summarizer to ensure that there was consistency and similarity in the comparisons. The results in Table I establish that the proposed summarizer is more *effective* and *efficient* than BERT Extractive Summarizer for call transcripts. The *punctuation-restoration-accuracy* scores for customer and agent summaries have also varied between 90-100% in all cases. It may also be noted here that the proposed summarizer is highly parameterized and provides more options than the BERT Extractive Summarizer.

**Customer Transcript - Extracted**

'she moved to a little one was thinking why not trying to understand crazy ones or players ability to change okay let me see and chop checks you all right no i mean you come up yeah i is there an a fortune was yes this is christmas cause see this is dont wanna be this yeah hes i will be all wanna actually calling because when i was thinking out sometimes problems the internet or something you know a wiring it im not sure if the internet itself is monthly somehow in knowing tina today nothing youre reading and its only if you have a clock is i see me really i never ask issue okay all over only okay all right really ok'

**Customer Transcript - Summary**

'I mean, you come up cause, see, do not want it to be like this. Because when i was thinking out sometimes problems, the internet, or something, you know, wiring it. I am not sure if the internet itself is monthly. Somehow, in knowing there is no alternative today, nothing you can read.'

**Agent Transcript - Extracted**

'im gonna get your name please right see i did not would you happen to have an account number no account came in alright thank you okay she yeah this is a wait this is dsl service do you do you have a tech go out today right im sorry i got this yeah i wont let me do this youre gonna get i got to get someone from dsl for you right ill get you in their call to be just a moment'

**Agent Transcript - Summary**

'Am I going to get your name, please, right? Would you happen to have an account number? Do you have a tech go out today? I got to get someone from DSL for you, right? I will get you in their call, to be just a moment.'

Fig. 4: Internet Service Disruption & Confirmation of Technician's Visit

**Customer Transcript - Extracted**

'i told the girl that i have to work in that but im on the back up you know you guys are scheduled to come out to see me tomorrow i told her the eleven and three you would not work one in five would actually work a little better i dont get off looking for i have been without phone and internet students got knows how long okay like i told her that was even feasible i know is boy you can call my cellphone its its my phone number and well cool for somebody doesnt even make sense that i should be without phone and internet service for the law shes telling me i get credit for the games i didnt use it but thats not the points i think a okay you know and neither channels good funny i said i need to see i get off work at four and im usually horrible for so that do that would make more sense than when im not moving to possibly yeah but they didnt change it yeah i guess right yes im okay i appreciate you know i need thank you no i said i just appreciate you coming back to update any yet i dont sure and thank you thank you'

**Customer Transcript - Summary**

'I do not get what I am looking for. I have been without phone and internet. God knows how long? Like i told her, that was even feasible. Does not even make sense that i should be without phone and internet service for that long.'

Fig. 5 (a): Customer Complaint about Phone & Internet Disruption

**Agent Transcript - Extracted**

'my apologies i see the account here i see all the notes take a look at this and find out whats going on for you said unfortunately it was you said three to eleven was not good i mean levin to three what they have noted on mask on speaking with today mission each and every do need to call you back and its what number can i call you back on right im gonna go and get this verify you one thing i just need to do for verification purposes i do have your account number available by chance the those last three digits thank you so much all right i see the ticket me take a look at what they have setup your let me see here still working on this year for a pre shared okay for right still working on this year for i see the order and so im trying to see we can find out whats going on its i mean i do see that levin to three still on there was what time during the day was good for you okay let me see mean if anything you think they would remove you know i give you me to the end of the day was gonna be best for you all right so i see what im gonna do is im gonna go to our dispatch center you said release you said after for thirty crack no let me see what i can find out for you were removed at and im gonna see if we can make this like maybe the last job day or make sure that the tech doesnt right after that for thirties when youre telling us out okay pushing a brief hold i really appreciate all be right back really for really its really through it no im still working on here are you okay hold on just another moment said it was more time no not a problem the least i can do for actually wanna see like just updated time but we do i do need to verify with our dispatch im just checking one more thing should only take a moment thank you and for are you really thats true router yeah are really and good really yeah im im still working out were just waiting a little bit longer and i should have an update for you thank you okay really it really it really okay for yeah thank you really a little better thats true the a and well thank you so much for waiting all apologize for the wait theres just a cure to get in touch and some just waiting for them to come to the line i just want it off you are you okay hold a few more minutes would you rather me call you back absolutely it do you want me to call you on the number youre calling me on sure well that number for make sure havent right yep thats the one the ticket yeah what do soon as i get an update from them which i think well be in actually its gotta be very soon forty'

Fig. 5 (b): Agent's confirmation of Technician's Visit

**Agent Transcript - Summary**

'I see the account here. I do have your account number available by chance. And i am going to see if we can make this, like maybe the last job today, or make sure that the tech does not go right after four thirty, when you are telling us you will be out, okay, pushing for a brief hold. I really appreciate the call, and i should have an update for you. Hold on for a few more minutes.'

**Summarizer Comparisons**

| Sample | Summarizer | Customer Rouge Score | Agent Rouge Score | Summarization Time (secs) |
|---|---|---|---|---|
| Sample-1 | Proposed Summarizer | 0.42 | 0.38 | 27 |
| | BERT Extractive Summarizer | 0.32 | 0.34 | 55.04 |
| Sample-2 | Proposed Summarizer | 0.49 | 0.44 | 22.81 |
| | BERT Extractive Summarizer | 0.35 | 0.35 | 45 |
| Sample-3 | Proposed Summarizer | 0.47 | 0.43 | 26.61 |
| | BERT Extractive Summarizer | 0.33 | 0.33 | 50.08 |
| Sample-4 | Proposed Summarizer | 0.5 | 0.45 | 341.05 |
| | BERT Extractive Summarizer | 0.37 | 0.36 | 796.59 |

Table I: Evaluation Metric scores for Extractive Summarizers from 4 samples of Call Transcripts

## VII. CONCLUSION

In this paper, we have presented an extractive summarization technique to address some of the challenges associated in general with call transcript summarization. We have combined *channel separation*, *topic modeling* and *sentence selection* with *punctuation restoration* to generate more readable call transcript summaries, in order to provide a better understanding of the customer complaints and the agent recommended solutions. This is perhaps the first summarizer that creates and evaluates multiple different types of topic models before selecting the most optimal one for summarization. We have provided a finer-grained similarity analysis, by using both *term-based similarities* for significant term extraction and *sentence-based similarities* for extractive summarization. This similarity analysis leverages both *Word2Vec* and *USE* based *embeddings* to exploit the semantic contents of words and sentences to determine their significance, uniqueness and relevance. The proposed extractive summarizer is the *only* one that restores *full punctuation* to the summaries generated from either ill-punctuated or unpunctuated original call transcripts, using a *novel BERT transformer based model*. We have introduced a *new metric* to evaluate the accuracy of the punctuation restoration in the resultant summaries. Finally, we have established the efficacy of the proposed summarizer through extensive evaluations and performance comparisons.

## CONFLICT OF INTEREST STATEMENT